\documentclass[runningheads]{llncs}
\usepackage[T1]{fontenc}
\usepackage{graphicx}
\usepackage{multirow}
\usepackage{amsmath}
\usepackage{makecell}
\usepackage{graphicx}
\usepackage{subcaption}
\usepackage{booktabs}
\usepackage{amssymb}
\usepackage{comment}
\usepackage[table]{xcolor}
\newcommand{\red}[1]{\textcolor{black}{#1}}
\begin{document}
\title{Towards Universal Khmer Text Recognition}
%
%

\author{Marry Kong\inst{1} \and Rina Buoy\inst{1,2} \and Sovisal Chenda\inst{1} \and Nguonly Taing\inst{1} \and Masakazu Iwamura\inst{2} \and Koichi Kise\inst{2}}
%
\authorrunning{Kong et al.}
%
\institute{Techo Startup Center, Ministry of Economy and Finance, Cambodia \and Osaka Metropolitan University, Japan}
\maketitle              
\begin{abstract}
Khmer is a low-resource language characterized by a complex script, presenting significant challenges for optical character recognition (OCR). While document printed text recognition has advanced because of available datasets, performance on other modalities, such as handwritten and scene text, remains limited by data scarcity. \red{Training modality-specific models for each modality does not allow cross-modality transfer learning, from which modalities with limited data could otherwise benefit. Moreover, deploying many modality-specific models results in significant memory overhead and requires error-prone routing each input image to the appropriate model. On the other hand, simply training on a combined dataset with a non-uniform data distribution across different modalities often leads to degraded performance on underrepresented modalities.} To address these, we propose a universal Khmer text recognition (UKTR) framework capable of handling diverse text modalities. Central to our method is a novel modality-aware adaptive feature selection (MAFS) technique designed to \red{adapt visual features according to a particular input image modality} and enhance recognition robustness across modalities. Extensive experiments demonstrate that our model achieves state-of-the-art (SoTA) performance. Furthermore, we introduce the first comprehensive benchmark for universal Khmer text recognition, which we release to the community to facilitate future research. Our datasets and models can be accessible via this gated repository\footnote{in review}.

\end{abstract}
\section{Introduction}\label{intro}

\begin{figure}
\centering

\begin{subfigure}[b]{0.45\textwidth}
    \centering
    {\includegraphics[width=\textwidth]{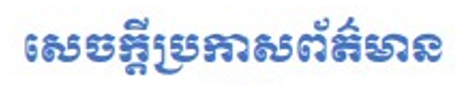}}
    \caption{document text}

\end{subfigure}
\hfill
\begin{subfigure}[b]{0.45\textwidth}
    \centering
    {\includegraphics[width=\textwidth]{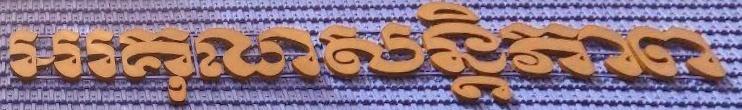}}
    \caption{scene text}

\end{subfigure}
\hfill
\begin{subfigure}[b]{0.45\textwidth}
    \centering
    {\includegraphics[width=\textwidth]{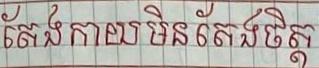}}
    \caption{handwritten text}

\end{subfigure}

\caption{Sample images from our new datasets.  } \label{image_samples_modalities}
\end{figure}

The development of Khmer text recognition has received significant attention over the past decade. However, many previous methods~\cite{buoy2022khmer,buoy2023toward,buoy2024language,buoy2025addressing} have primarily been designed to operate only on printed document modalities. This is mainly because it is relatively straightforward to generate synthetic printed document training data in sufficient quantity and quality. In contrast, generating high-fidelity synthetic data for other modalities, such as handwritten and scene documents as shown in Figure~\ref{image_samples_modalities}, remains technically challenging. Consequently, progress in Khmer scene text recognition, and particularly in handwritten text recognition, has been severely limited by the lack of high-quality, real-world training and benchmark datasets.

Due to the non-uniform distribution of Khmer text recognition data across different modalities (more printed document datasets and less scene and handwritten datasets), most previous methods~\cite{nom2024khmerst,valy2018character,nom2025cross} have been specifically designed and trained to recognize a single text modality. As a result, cross-modality learning is either not exploited at all or, at best, limited to pre-trained weight initialization~\cite{nom2025cross}. \red{Moreover, deploying multiple modality-specific models results in significant memory overhead and requires routing each input image to the appropriate model, making it unsuitable or error-prone for an end-to-end OCR pipeline.} Conversely, training a unified Khmer text recognition model on a combined dataset spanning all modalities with unequal proportions often leads to degraded performance on underrepresented modalities (i.e., handwritten and scene texts).

To achieve robust recognition performance across all text modalities, this paper proposes a universal Khmer text recognition framework (UKTR) capable of handling diverse text modalities. This is achieved through a novel modality-aware adaptive feature selection (MAFS) technique, which dynamically selects the most relevant visual features required for accurately recognizing text across different modalities. \red{In addition, the proposed UKTR framework employs a joint connectionist temporal classification (CTC)~\cite{graves2006connectionist} and Transformers-based decoder~\cite{vaswani_attention_2017} to generate text in a non-autoregressive (i.e., generating all tokens in parallel) and autoregressive manner (i.e., generating one token at a time), respectively. The two decoders present a latency-accuracy trade-off within the same model, with the CTC decoder being faster but less accurate and the Transformer-based decoder being slower but more accurate.}

To facilitate robust training and evaluation of the proposed UKTR framework, we construct the first joint general Khmer scene and handwritten text datasets and benchmarks, which will serve as a valuable resource for the research community beyond this work. Experimental results demonstrate that the proposed UKTR model achieves state-of-the-art (SoTA) performance on this new multi-modality and existing benchmarks.

 Our contributions can be summarized as follows:
\begin{enumerate}
    \item We propose the UKTR framework that recognizes Khmer text across printed, scene, and handwritten modalities, robustly. The proposed framework is based on a novel MAFS technique that adaptively selects the most relevant visual features required for accurate text recognition across different modalities.

    \item Our UKTR model supports both non-autoregressive and autoregressive text generation, enabling a flexible trade-off between latency and accuracy during inference.
    
    \item We construct the first joint general Khmer scene and handwritten text datasets and benchmarks.

    \item Our proposed UKTR model achieves SoTA performance on the new multi-modality and existing benchmarks.
\end{enumerate}


\section{Related Work}\label{relatedwork}

\begin{figure}[t]
\centering

\centering
    {\includegraphics[width=\hsize]{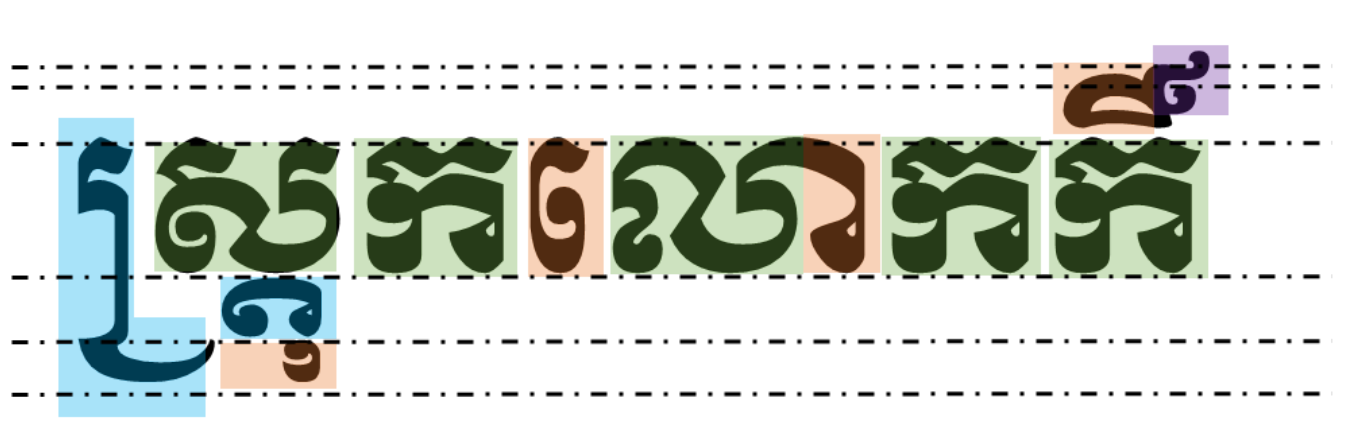}}
    \caption{Sample Khmer text layout (with permission~\cite{buoy2025addressing}.)  Blue: consonant subscript. Green: base consonant. Orange: dependent vowel. Purple: diacritic. Best viewed in color. }

\label{khmer_text_layout}
\end{figure}

Khmer is a complex abugida writing system with a large inventory of characters, including consonants (base and subscript forms), dependent vowels, independent vowels, and diacritics~\cite{buoy2023toward}, as illustrated in Figure~\ref{khmer_text_layout}. Khmer text is written from left to right with optional spaces. Characters are often stacked by attaching subscripts to a base consonant, forming character clusters with ligatures~\cite{buoy2023toward}. Such complex stacking structures pose significant challenges for accurate character recognition.

Compared with Latin-based languages, Khmer is a low-resource and low-attention language, with limited research and publicly available datasets. Nonetheless, over the past decade, deep learning-based approaches have been proposed to advance Khmer text recognition, particularly for the printed text modality. Prior to the deep learning era, Khmer OCR methods primarily focused on isolated character recognition, employing wavelet-based feature descriptors~\cite{chey2006khmer} and support vector machine (SVM) classifiers~\cite{sok2014support}. The performance and generalization capabilities of these classical machine learning approaches were significantly limited.

For printed documents and synthetic scene text modalities, Buoy et al.~\cite{buoy2022khmer} introduced a sequence-to-sequence (seq2seq) Khmer textline recognition framework with an attention mechanism, consisting of a convolutional visual feature extractor and an autoregressive decoder based on a gated recurrent unit (GRU) for character decoding. Buoy et al.~\cite{buoy2023toward} extended this work by employing a 2D visual feature extractor to enhance feature representation and a Transformers-based decoder~\cite{vaswani_attention_2017} to improve character decoding performance. Their models were trained on large-scale synthetically generated document and scene text datasets. \red{Similarly, Keo et al.~\cite{keo2024state} trained both CTC-based and attention-based Khmer text recognition models using synthetic word-level training images.} Since Khmer textlines can be very long and the decoding process is prone to attention misalignment, Buoy et al.~\cite{buoy2025addressing} proposed a new recognition approach based on a partially autoregressive (PAR) decoder. While previous methods treat individual characters as atomic units, the authors introduced a non-autoregressive framework that generates text at the character cluster level, significantly reducing decoding time while maintaining recognition accuracy. In addition, off-the-shelf OCR systems such as Surya-OCR~\cite{paruchuri2025surya}, Tesseract-OCR~\cite{tesseract_github}, and emerging multimodal models, including DeepSeek-OCR (v1 and v2)~\cite{wei2025deepseek,wei2026deepseek}, are also capable of recognizing Khmer text for certain fonts. Among these, Tesseract-OCR has served as a strong baseline for Khmer OCR.

For the real scene text modality, Nom et al.~\cite{nom2024khmerst} published the first real-world Khmer scene text dataset, named KhmerST, comprising images captured using smartphone cameras. The dataset was designed to support both text detection and recognition tasks. For text recognition, they fine-tuned a Transformers-based OCR (TrOCR)~\cite{li2023trocr} model, achieving superior performance compared to Tesseract-OCR on Khmer text. Furthermore, Nom et al.~\cite{nom2025wildkhmerst} introduced an expanded version of the dataset, called WildKhmerST, which contains more diverse and challenging real-world samples. Nom et al.~\cite{nom2025cross} subsequently investigated cross-lingual learning strategies for Khmer text recognition by exploring various pre-trained visual encoders and language decoders, demonstrating significant improvements in both recognition accuracy and generalization capability.

For the handwritten text modality, previous research has been largely limited to isolated character recognition~\cite{annanurov2021compact} and short-word recognition~\cite{valy2018character}, primarily focusing on historical palm-leaf manuscripts. \red{Recently, Ngin et al.~\cite{ngin2026khmerwriterid} constructed a Khmer online handwriting dataset that is used for Khmer writer verification purposes.} Consequently, Khmer handwritten text recognition, particularly for modern Khmer, remains relatively underexplored.

\red{In summary, current Khmer OCR research primarily focuses on developing modality-specific models. However, training separate models for each modality prevents cross-modality transfer learning, which could otherwise benefit modalities with limited data. In an end-to-end OCR pipeline, deploying separate models for each input modality results in significant memory overhead and increases the likelihood of errors, as each input must be routed to the appropriate model.} Therefore, in this paper, we propose a holistic approach to Khmer text recognition, aiming to accurately recognize Khmer text across all modalities. We address this challenge from both the dataset and model architecture perspectives.

\section{Methodology}\label{methodology} 

\subsection{Proposed Universal Khmer Text Recognition Framework (UKTR)}\label{proposed_method} 

\begin{figure}[t]
\centering
\includegraphics[width=\textwidth]{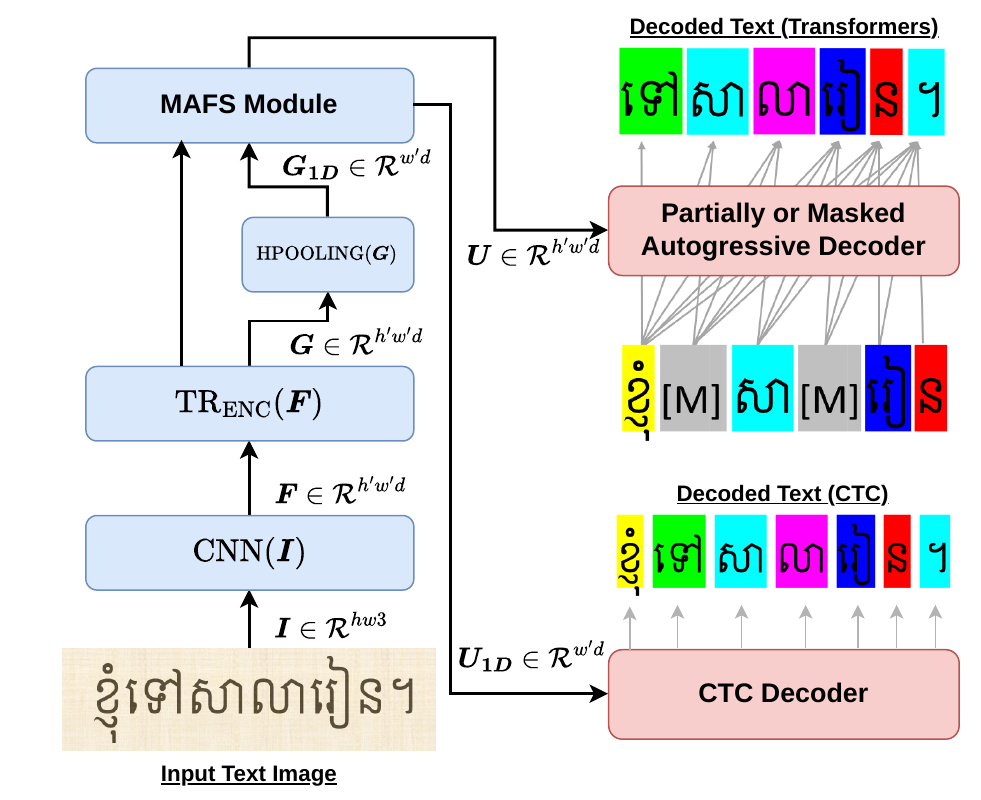}
\caption{The proposed UKTR framework.} \label{overall_architecture}
\end{figure}

The proposed UKTR framework, as shown in Figure~\ref{overall_architecture}, is designed to robustly recognize Khmer text across multiple modalities. It supports text generation in both non-autoregressive (NAR) and autoregressive (AR) modes. The architecture comprises a base convolutional visual encoder, a modality-aware adaptive feature selector, a CTC-based NAR decoder, and a Transformers-based AR decoder.

\subsection{Visual Encoder}\label{visual_encoder} 

The visual encoder is designed to extract the essential visual features required for accurate character recognition. Following~\cite{buoy2025addressing}, our visual encoder consists of a base convolutional network for visual feature extraction, followed by a Transformers-based encoder to capture sequential dependencies.

Mathematically, the visual encoder can be expressed as

\noindent
\begin{align}\label{eq:1}
\boldsymbol{F} &=\operatorname{\mathrm{CNN}}(\boldsymbol{I})\\
\boldsymbol{G} &=\operatorname{\mathrm{TR}_{ENC}}(\boldsymbol{F})\\
\boldsymbol{G_{1D}} &=\operatorname{\mathrm{HPOOLING}}(\boldsymbol{G}),
\end{align}
where $\operatorname{\mathrm{CNN}}$ and $\operatorname{\mathrm{TR}_{ENC}}$ are the base convolutional feature extractor and the Transformers encoder network, respectively.  $\boldsymbol{I}$ is the input RGB image, $\boldsymbol{F}$ and $\boldsymbol{G}$  are 2D feature maps with a depth of $d$. Since the CTC decoder expects 1D feature maps, $\operatorname{\mathrm{HPOOLING}}$ is used to average the 2D feature maps over the height dimension.

The base $\operatorname{\mathrm{CNN}}$ feature extractor consists of six sequential ResNet blocks with increasing channel dimensions from 32 to 512, where each block is formed by 1×1 and/or 3×3 convolutions repeated multiple times. Spatial downsampling by a factor of $(2,2)$ is applied in ResNet-Blocks 1 and 4, while the remaining blocks preserve the spatial resolution. The $\operatorname{\mathrm{TR}_{ENC}}$ network uses an embedding dimension ($d$) of 512, with three stacked encoder layers, eight attention heads, a dropout rate of 0.1, and a feed-forward network dimension of 2048 ($4\times d$). The visual encoder's model size and complexity are given in Table~\ref{encoder_params_flops}.

\begin{table}[t]
\centering
\caption{The visual encoder's complexity and model size. Params: model parameters. Flops: floating operations on an input of $32\times116$ pixels. }\label{encoder_params_flops}
\begin{tabular}{|lrr}
\hline
\textbf{Name} &  \textbf{Params.} &\textbf{Flops} \\
\hline
ResNet backbone (visual features) & 13.0M & 3.2G \\
Transformers encoder (sequential features)& 9.5M & 2.2G  \\
Total & 22.5M & 5.4G \\
\hline

\hline
\end{tabular}
\end{table}

\subsection{Modality-Aware Adaptive Feature Selector (MAFS)}\label{feature_selector} 

To enable the proposed framework to robustly recognize Khmer text across different modalities, we introduce a modality-aware adaptive feature selector, which consists of three components: modality router, modality adapter, and multi-modality aggregator. Mathematically, the modality-aware adaptive feature selector can be expressed as

\noindent
\begin{align}\label{eq:1}
\boldsymbol{z} &=\operatorname{\mathrm{GPOOLING}}(\boldsymbol{G})\\
\boldsymbol{r} &=\operatorname{\mathrm{ROUTER}}(\boldsymbol{z})\\
\boldsymbol{H} &=\operatorname{\mathrm{ADAPTER}}(\boldsymbol{G})\\
\boldsymbol{H_{1D}} &=\operatorname{\mathrm{ADAPTER}}(\boldsymbol{G_{1D}})\\
\boldsymbol{U} &=H\cdot \boldsymbol{r}\\
\boldsymbol{U_{1D}} &=H_{1D}\cdot \boldsymbol{r},
\end{align}
 where $\operatorname{\mathrm{GPOOLING}}$ is an average global pooling operator, which takes $\boldsymbol{G}$ and returns an average vector $\boldsymbol{z} \in R^d$. $\operatorname{\mathrm{ROUTER}}$ is the router network, whick takes $\boldsymbol{z}$ and returns a probability vector $\boldsymbol{r} \in R^n$ over $n$ text modalities (default: 5). $\operatorname{\mathrm{ADAPTER}}$ is a feature projection network with a default project dimension ($p$) of 128, which takes $\boldsymbol{G}$ and $\boldsymbol{G_{1D}}$ and returns  $\boldsymbol{H}$ and $\boldsymbol{H_{1D}}$, respectively. $\boldsymbol{U_{1D}}$ and $\boldsymbol{U}$ are the modality-aware adaptive features, which are marginalized over all modality sources, for the CTC and Transformers decoders, respectively.

\begin{figure}[t]
\centering
\begin{subfigure}[b]{0.45\textwidth}
    \centering
    {\includegraphics[width=\textwidth]{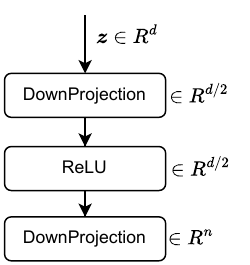}}
    \caption{The $\operatorname{\mathrm{ROUTER}}$ network. }
\end{subfigure}
\hfill
\begin{subfigure}[b]{0.48\textwidth}
    \centering
    {\includegraphics[width=\textwidth]{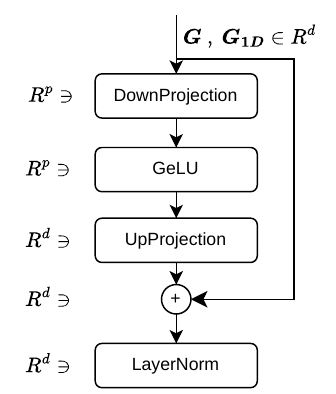}}
    \caption{The $\operatorname{\mathrm{ADAPTER}}$ network.}
\end{subfigure}

\caption{The designs of the $\operatorname{\mathrm{ROUTER}}$ and $\operatorname{\mathrm{ADAPTER}}$  networks.  } \label{router_adapter}
\end{figure}

\red{As shown in Figure~\ref{router_adapter}, the role of the $\operatorname{\mathrm{ADAPTER}}$  is to adapt visual features to different modalities (five adapters are used by default). Since the input modality is not known in advance and is not assumed to be available, the $\operatorname{\mathrm{ROUTER}}$  estimates the probability distribution over all modalities. Based on this probability distribution, the adapted features are combined accordingly. This design eliminates the need for prior knowledge of the image modality and is therefore well suited for real-world deployment.}

\red{Although there are three discrete modalities, document, scene, and handwritten, real images rarely belong to only one modality. Instead, they can be represented within a simplex spanning these three modalities. For example, a text image may be both handwritten and scene simultaneously. By default, we set this hyperparameter ($n$) to five.}


\subsection{Text Decoders}\label{decoders} 
To support both non-autoregressive (NAR) and autoregressive (AR) decoding within the same UKTR model, we incorporate both CTC and Transformers-based decoders. The CTC decoder aligns the $\boldsymbol{U_{1D}}$ feature maps with the target sequence $\boldsymbol{Y}$ by minimizing a CTC loss ($\boldsymbol{l_{CTC}}$) between the predicted token distributions $\hat{\boldsymbol{Y}}_{CTC}$, where $y_i \in \mathbb{R}^c$ ($c$ is the number of unique tokens), and $\boldsymbol{Y}$.

The Transformers decoder converts the $\boldsymbol{H}$ feature maps along with the context vector (i.e., $\boldsymbol{Y}$ shifted one token to the right) into $\hat{\boldsymbol{Y}}_{TR}$, where $y_i \in \mathbb{R}^c$. It is trained by minimizing the cross-entropy loss ($\boldsymbol{l_{TR}}$) between $\hat{\boldsymbol{Y}}_{TR}$ and $\boldsymbol{Y}$. The Transformers decoder consists of three stacked layers with an embedding dimension ($d$) of 512, 8 attention heads, a feed-forward network dimension of 2048, and a dropout rate of 0.1. Following Buoy et al.~\cite{buoy2025addressing}, a masking ratio of 30\% is applied. The Transformers decoder's complexity and model size are given in Table~\ref{decoder_complexity_param}.

As for the tokenizer, we extended the same Khmer character cluster tokenizer~\cite{buoy2025addressing} by adding case-sensitive English characters, numbers, and symbols.  The number of unique tokens ($c$) is 11,899. Mathematically, the total loss can be expressed as

\noindent
\begin{align}\label{eq:1}
\boldsymbol{l_{Total}} =\boldsymbol{l_{CTC}} + \boldsymbol{l_{TR}}.
\end{align}

\begin{table}[t]
\centering
\caption{The Transformers decoder's complexity and model size. Flops: floating operations on an input of $32\times116$ pixels.}\label{decoder_complexity_param}
\begin{tabular}{lrr}
\hline
\textbf{Name} &  \textbf{Params.} &\textbf{Flops} \\
\hline
Transformers decoder (AR decoding) & 24.08M & 0.82G \\
\hline

\hline
\end{tabular}
\end{table}

\section{Datasets}\label{experiments} 
Previous methods~\cite{buoy2022khmer,buoy2023toward,buoy2024language,buoy2025addressing} have primarily relied on synthetic training data, as it is technically convenient to generate large-scale datasets. In contrast, training data for scene and handwritten text modalities remain scarce for the Khmer script. In the following subsections, we first describe the existing publicly available datasets, \red{followed by a description of our newly developed datasets designed to address the gaps in Khmer scene and handwritten text data.}

\subsubsection{Existing Datasets}\label{khmer_dataset_existing} Here are the existing available Khmer text recognition datasets: 
\begin{itemize}

    \item \textbf{Buoy et al.~\cite{buoy2023toward}'s Synthetic Document and Scene Textline Data}: Buoy et al.~\cite{buoy2023toward} synthetically generated approximately 2.8 million document and scene textline images, comprising 1.5 million document images and 1.3 million scene images, using 11 commonly used Khmer fonts.
    \item \textbf{KHOB}~\cite{khob}: KHOB is a collection of Khmer textline images manually cropped from scanned low-resolution Khmer PDF documents. The processed dataset~\cite{buoy2025addressing} contains 1,318 textline samples.
    \item \textbf{SynthText~\cite{Seanghay2024SynthKhmer10k}}: 70,000 textline images were extracted from 10,000 synthetically generated Khmer identity card images.
    \item \textbf{Long et al.~\cite{long2022towards}'s HierText Data}:  This dataset includes training (8,281 images), validation (1,724 images), and test (1,634 images) sets, with annotations for both Latin printed and handwritten textlines and words. After cropping into textlines and removing vertical texts from the training and validation sets, the dataset contains approximately 518,726 images.
    \item  \textbf{Nom et al.~\cite{nom2024khmerst}'s KhmerST}: This dataset contains real Khmer scene text images, mostly logos and billboards, captured in natural scenes (both indoor and outdoor). It includes a mixture of short words and long text 
    lines. The dataset originally comprised 1,544 manually annotated images, yielding a total of 3,237 cropped text images. The processed KhmerST dataset~\cite{buoy2025addressing}  contains 3,022 text images. The dataset is particularly challenging due to complex imaging conditions and diverse artistic fonts.
    \item  \textbf{Nom et al.~\cite{nom2025wildkhmerst}'s WildKhmerST}: This dataset represents an expanded version of KhmerST, containing 29,601 annotated textlines from 10,000 images captured in diverse public locations across Cambodia, such as streets, signboards, supermarkets, and commercial areas. It includes challenging text samples, including artistic, blurred, low-light, curved, and occluded text, as well as text embedded in complex backgrounds.
    \item \textbf{Soy~\cite{SoyVitou2025KhmerHandwritten42k}'s Khmer Handwritten Data (KH)}: This dataset contains approximately 4.2k handwritten and printed Khmer and Latin textline images. It is divided into training (3,991 images) and evaluation (211 images) sets.
\end{itemize}

\begin{figure}[t]
\centering
\begin{subfigure}[b]{0.45\textwidth}
    \centering
    {\includegraphics[width=\textwidth]{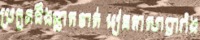}}
    \caption{scene (synthetic)}
\end{subfigure}
\hfill
\begin{subfigure}[b]{0.45\textwidth}
    \centering
    {\includegraphics[width=\textwidth]{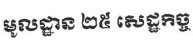}}
    \caption{document (synthetic)}

\end{subfigure}
\hfill
\begin{subfigure}[b]{0.45\textwidth}
    \centering
    {\includegraphics[width=\textwidth]{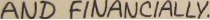}}
    \caption{scene (real, Latin)}

\end{subfigure}
\hfill
\begin{subfigure}[b]{0.45\textwidth}
    \centering
    {\includegraphics[width=\textwidth]{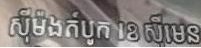}}
    \caption{scene (real)}

\end{subfigure}
\hfill
\begin{subfigure}[b]{0.45\textwidth}
    \centering
    {\includegraphics[width=\textwidth]{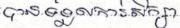}}
    \caption{handwritten}

\end{subfigure}

\caption{Sample cropped images from the existing datasets. (a) Buoy et al.~\cite{buoy2023toward}. (b) KHOB. (c) HierText. (d) KhmerST.   } \label{sample_images_existing}
\end{figure}

The summary of the existing datasets and sample preview images are provided in Table~\ref{tab:dataset_summary} and Figure~\ref{sample_images_existing}, respectively.

\subsubsection{New Datasets in this Study}\label{khmer_dataset_new}
\begin{figure}[t]
\centering

\begin{subfigure}[b]{0.48\textwidth}
    \centering
    {\includegraphics[width=\textwidth]{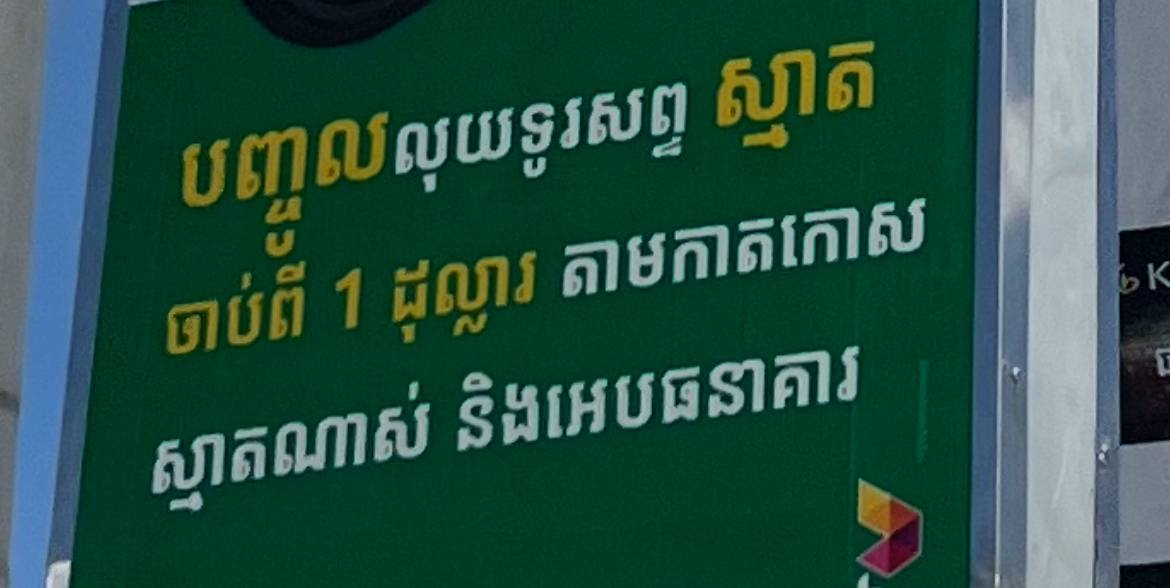}}
    \caption{WildKhmerST}

\end{subfigure}
\hfill
\begin{subfigure}[b]{0.48\textwidth}
    \centering
    {\includegraphics[width=\textwidth]{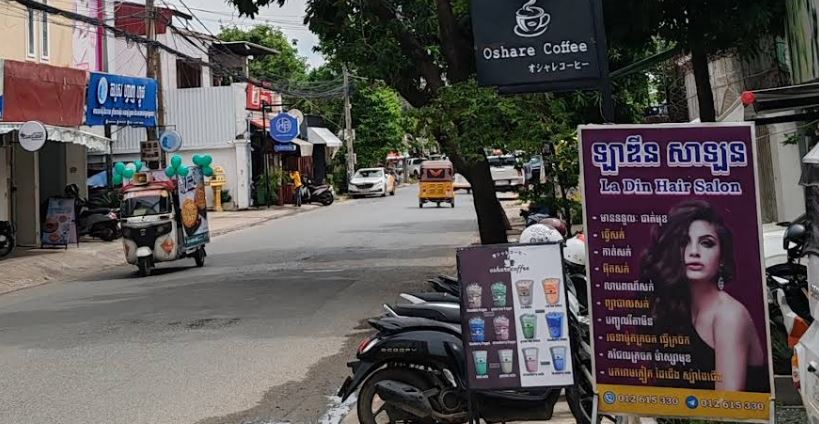}}
    \caption{GKST (our)}

\end{subfigure}

\caption{Sample images from the WildKhmerST dataset and our GKST dataset.  } \label{wildkhmerstvsgkst}
\end{figure}

\begin{figure}[t]
\centering

\begin{subfigure}[b]{0.4\textwidth}
    \centering
    {\includegraphics[width=\textwidth]{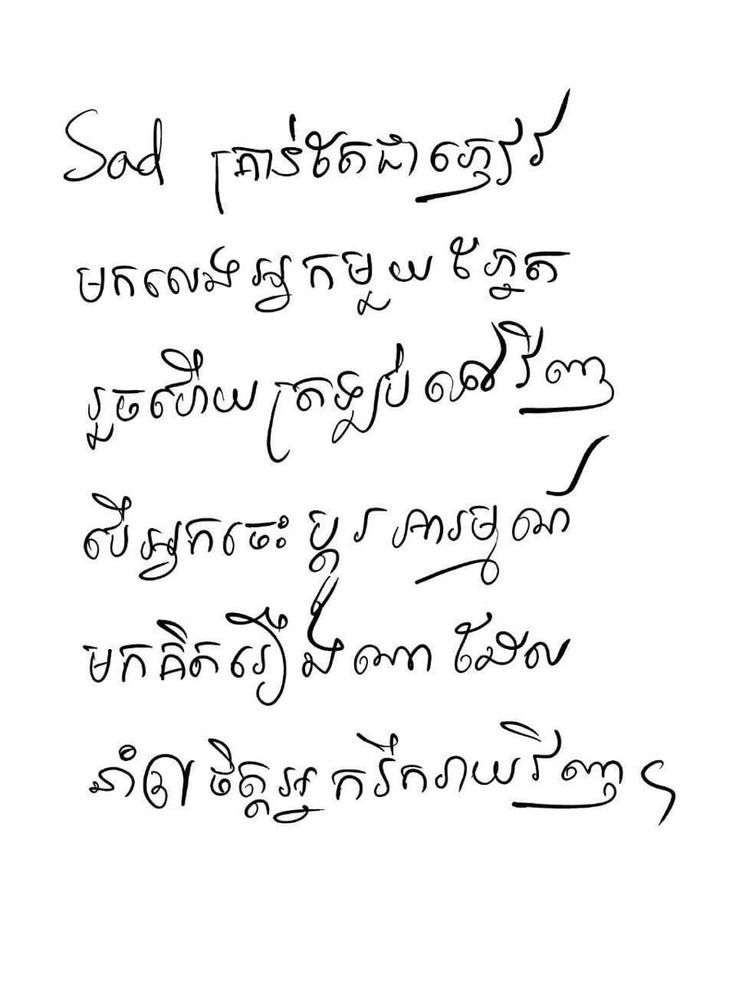}}
    \caption{handwritten text on paper}

\end{subfigure}
\hfill
\begin{subfigure}[b]{0.5\textwidth}
    \centering
    {\includegraphics[width=\textwidth]{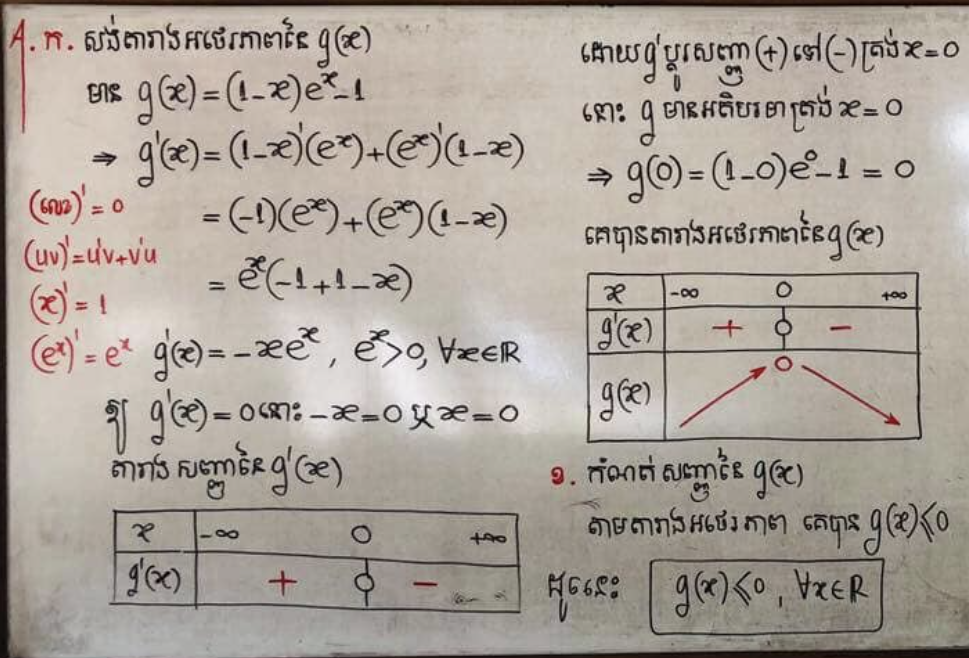}}
    \caption{handwritten text on whiteboard}

\end{subfigure}

\caption{Sample images from our KHT dataset.  } \label{our_kht_samples}
\end{figure}

: To address the limited availability of Khmer scene and handwritten text images, we introduce two new datasets for these modalities, as described below:
\begin{itemize}
    \item \textbf{General Khmer Scene Text Data (GKST):} We manually annotated 4,221 Khmer scene text images. Similar to the KhmerST and WildKhmerST datasets, these images were captured using smartphone cameras under general indoor and outdoor conditions, exhibiting natural distortions and perspective transformations. \red{owever, a key distinction is that we started with general scene images and annotated the text within them. In contrast, the KhmerST and WildKhmerST datasets contain focused, close-up text images, as shown in Figure~\ref{wildkhmerstvsgkst}.}  The dataset is divided into train (4,009 images) and evaluation (212 images) sets.
    \item \textbf{General Khmer Handwritten Text Data (KHT):} Similarly, we manually annotated 14,168 Khmer handwritten text images from diverse sources, including handwritten birth certificates, exam papers, and notes, as shown in Figure~\ref{our_kht_samples}. The dataset is divided into training (13,457 images) and evaluation (711 images) sets.
\end{itemize}

\begin{figure}[t]
\centering

\begin{subfigure}[b]{0.48\textwidth}
    \centering
    {\includegraphics[width=\textwidth]{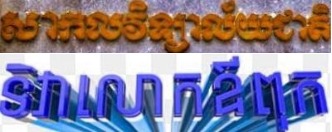}}
    \caption{scene (real)}

\end{subfigure}
\hfill
\begin{subfigure}[b]{0.48\textwidth}
    \centering
    {\includegraphics[width=\textwidth]{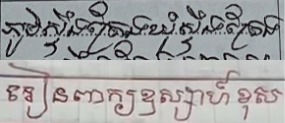}}
    \caption{handwritten (real)}

\end{subfigure}

\caption{Sample cropped images from our new datasets.  } \label{sample_images_new}
\end{figure}

The summary of the new datasets and sample preview images are provided in Table~\ref{tab:dataset_summary} and Figure~\ref{sample_images_new}, respectively.

\subsubsection{Document, Scene \& Handwritten, and Evaluation Datasets}:

We categorize the above datasets into three groups according to their primary modality as follows:
\begin{itemize}
    \item \textbf{Document (D)}: The datasets in this group are primarily synthetic but large-scale, with the exception of the Latin HierText dataset. This group includes Buoy et al.~\cite{buoy2023toward}, SynthText~\cite{Seanghay2024SynthKhmer10k}, and HierText~\cite{long2022towards}. The objective of this group is to learn visual representations of both Khmer and Latin scripts from large-scale printed document text data.

    \item  \textbf{Scene \& Handwritten (S\&H)}: The datasets in this group consist of real data sources, primarily comprising scene and handwritten text. The group includes WildKhmerST~\cite{nom2025wildkhmerst}, the training sets of Soy~\cite{SoyVitou2025KhmerHandwritten42k}, and our newly collected datasets.

    \item \textbf{Evaluation}: To maintain consistency with the previous methods, the evaluation group includes the KhmerST~\cite{nom2024khmerst} dataset (S), the KHOB~\cite{khob} dataset (D), the evaluation sets of KH~\cite{SoyVitou2025KhmerHandwritten42k} (D\&H), and our newly introduced GKHST and KHT datasets (S\&H).
\end{itemize}

\begin{table*}[t]
\centering
\caption{Summary of datasets, sizes, usage, and modalities.}
\label{tab:dataset_summary}
\begin{tabular}{lcccccccc}
\toprule
Dataset & Train & Eval & Training (D) & Adapting (S\&H) & Eval & Doc & Scene & Handwritten \\
\midrule
Buoy et al.~\cite{buoy2023toward} & 2.8M & -- & \checkmark & -- & -- & \checkmark & \checkmark & -- \\
KHOB~\cite{khob} & -- & 1,318 & -- & \checkmark & \checkmark & \checkmark & -- & -- \\
SynthText~\cite{Seanghay2024SynthKhmer10k}& 70k & -- & \checkmark & -- & -- & \checkmark & -- & -- \\
HierText~\cite{long2022towards} & 518,726 & -- & \checkmark & -- & -- & -- & \checkmark & \checkmark \\
KhmerST~\cite{nom2024khmerst} & -- & 3,022 & -- & \checkmark & \checkmark & -- & \checkmark & -- \\
WildKhmerST~\cite{nom2025wildkhmerst} & 29,601 & -- & -- & \checkmark & -- & -- & \checkmark & -- \\
KH~\cite{SoyVitou2025KhmerHandwritten42k}& 3,991 & 211 & -- & \checkmark & \checkmark & \checkmark &  -- & \checkmark \\
GKST (ours) & 4,009 & 212 & -- & \checkmark & \checkmark & -- & \checkmark & -- \\
KHT (ours) & 13,457 & 711 & -- & \checkmark & \checkmark & -- & -- & \checkmark \\
\bottomrule
\end{tabular}
\end{table*}

\section{Experiments, Results, and Discussion}\label{Experiments} 

\subsection{Experimental Setup} \red{The UKTR models were first trained on the large-scale document (D group) datasets to learn robust visual representations of Khmer and Latin characters. During this phase, a cyclic learning rate schedule was employed, with a minimum value of $10^{-5}$ and a maximum of $10^{-4}$. This training phase lasted five full epochs with a batch size of 32 images.}

\red{The trained UKTR models then underwent a modality-adapting training phase on the scene \& document (S\&H group) datasets across different modalities for 50 epochs. During this phase, a lower cyclic learning rate schedule was employed, with a minimum value of $10^{-6}$ and a maximum of $10^{-5}$. In both phases, the Adam optimizer was used, with a gradient clipping value of 50.}

\red{To prevent the models from underperforming on printed document text images during the modality-adapting training phase, we sampled an equal number of document images and mixed them with the S\&H images. This mixing strategy, together with the proposed MAFS module, enables the models to maintain robustness across all text modalities.}

\subsection{Results and Discussion}\label{result} 
In terms of evaluation metrics, only character error rate (CER) is reported for evaluating Khmer text recognition performance in the subsequent analyses, as Khmer is an unsegmented script (i.e., it has no explicit word boundaries).
\subsubsection{Recognition Performance of the UKTR models }\label{khmer_optimal_style} Table~\ref{tab:reg_accuracy} presents the recognition accuracy comparisons of our UKTR models against previous methods across evaluation datasets of different modalities. The results show that our models (\textbf{UKTR (D + S\&H)}) outperform previous methods by a significant margin on all evaluation datasets except KHOB. Specifically, our best model using the Transformers decoder achieved CERs of 2.37\%, 2.19\%, 4.11\%, 3.34\%, and 6.10\% on the KHOB, KhmerST, KH, GKHST, and KHT datasets, respectively.

On the KHOB dataset (D), our best model marginally underperformed compared to Buoy et al.~\cite{buoy2025addressing}, which achieved a CER of 2.13\%. This is because their model, with its dedicated Transformers decoder, was specifically optimized and trained for recognizing printed document texts. In contrast, our UKTR models, employing both joint CTC and Transformers decoders, were trained and optimized to recognize Khmer text across multiple modalities.

For our UKTR models, comparing the Transformers decoder with the CTC decoder shows that the former moderately outperforms the latter. Specifically, the CER improvements achieved by the Transformers decoder are 0.90\%, 0.83\%, 1.78\%, 1.07\%, and 3.42\% on the KHOB, KhmerST, KH, GKHST, and KHT datasets, respectively. These results highlight the important role of language modeling enabled by the Transformers decoder in Khmer character recognition. However, these gains come at the cost of increased recognition latency, since the CTC decoder generates all tokens simultaneously, whereas the Transformers decoder generates tokens sequentially, one at a time.

For our UKTR models, comparing the general training phase (i.e.,\textbf{UKTR (D Only)}) using the synthetic document datasets only with the modality-adapting training phase (\textbf{UKTR (D + S\&H)}) using both real scene and handwritten datasets shows that the modality-adapted models achieve significantly lower CERs across all evaluation datasets. This demonstrates that our modality-adapting strategy successfully retains general training capabilities (for printed document recognition) while simultaneously acquiring new capabilities for additional modalities (handwritten and scene text). This is facilitated by the proposed MAFS module, which is modality-aware and can adaptively select visual features based on the specific text modality. The ablation study of the proposed MAFS module will be discussed in the next subsection.

In summary, our UKTR models achieve SoTA recognition performance across all modalities, including document, scene, and handwritten texts. The same UKTR models can be configured to perform both autoregressive and non-autoregressive decoding.

\begin{table*}[t]
\centering
\caption{Character error rates (CER) on the evaluation datasets of different modalities (D,S,\&H). * : model-specific tokenizer. Char.: character. KCC: Khmer character cluster. \textbf{Bold}: best. \emph{Italic}: second best.}
\label{tab:reg_accuracy}
\begin{tabular}{lcccccccc}
\toprule
 & Dec. & Tok. & KHOB(D) & KhmerST(S) & KH(D,H) & GKST(S) & KHT(H) \\
\midrule
Tesseract-OCR~\cite{tesseract_github} & CTC & Char. & 9.19 & 40.96 & -- & -- & -- \\
Surya-OCR~\cite{paruchuri2025surya} & Tr. & * & 17.69 & 43.21 & -- & -- & -- \\
Buoy et al.~\cite{buoy2023toward} & Tr. & Char. & 3.03 & -- & -- & -- & -- \\

Buoy et al.~\cite{buoy2024language} & CTC & KCC & 2.33 & -- & -- & -- & -- \\
Nom et al.~\cite{nom2024khmerst} & Tr. & * & -- & 17.00 & -- & -- & -- \\
Soy et al.~\cite{SoyVitou2025KhmerHandwritten42k} & Tr. & * & -- & -- & 17.00 & -- & -- \\
Buoy et al.~\cite{buoy2025addressing} & Tr. & KCC& \textbf{2.13} & 7.01 & -- & -- & -- \\
\midrule
\textbf{UKTR (D Only)} & CTC & KCC &  2.70 & 10.06 & 32.96 & 10.50 & 45.90\\
\textbf{UKTR (D Only)} & Tr. & KCC &  2.55 & 14.34 & 36.30 & 15.26 & 52.52\\
\textbf{UKTR (D + S\&H)} & CTC & KCC &  2.46 & \emph{3.02} & \emph{5.89} & \emph{4.41} & \emph{9.52}\\
\textbf{UKTR (D + S\&H)} & Tr. & KCC &  \emph{2.37} & \textbf{2.19} & \textbf{4.11} & \textbf{3.34} & \textbf{6.10}\\
\bottomrule
\end{tabular}
\end{table*}

\subsubsection{Ablation Study of the MAFS Module}\label{japanese_results} To evaluate the robustness of the proposed MAFS module, we established a baseline model in which the module was removed. Table~\ref{tab:mafs} compares the CERs of models with and without the MAFS module. The results show that removing the MAFS module leads to a significant degradation in modality-adapting performance across all text modalities, for both decoder types. This occurs because, during modality-adapting training, the model without the MAFS module struggles to learn the relevant visual features required to accurately recognize Khmer text for a given input modality. Consequently, the model fails to retain its general training capabilities (for printed document text recognition) and is unable to acquire new capabilities for scene and handwritten text modalities.
\begin{table*}[t]
\centering
\caption{Character error rates (CER) w/ and w/o the MAFS module. * : model-specific tokenizer. Char.: character. KCC: Khmer character cluster. \textbf{Bold}: best. \emph{Italic}: second best.}
\label{tab:mafs}
\begin{tabular}{lcccccccc}
\toprule
 & Decoder & Tokenizer & KHOB & KhmerST & KH & GKST & KHT \\

\midrule
\textbf{UKTR w/o MAFS (D + S\&H)} & CTC & KCC &  3.85 & 4.30 & 8.17 & 6.15 & 11.89\\
\textbf{UKTR w/o MAFS (D + S\&H)} & Tr. & KCC &  3.93 & 5.48 & 5.84 & 5.39 & \emph{7.66}\\
\textbf{UKTR  w/ MAFS (D + S\&H)} & CTC & KCC &  \emph{2.46} & \emph{3.02} & \emph{5.89} & \emph{4.41} & 9.52\\
\textbf{UKTR  w/ MAFS (D + S\&H)} & Tr. & KCC &  \textbf{2.37} & \textbf{2.19} & \textbf{4.11} & \textbf{3.34} & \textbf{6.10}\\
\bottomrule
\end{tabular}
\end{table*}

\subsubsection{Effect of the Number of Modality Sources }\label{japanese_results} 
The $\operatorname{\mathrm{ROUTER}}$ network estimates a probability distribution over modality sources ($n=5$). To evaluate the impact of this choice, we also considered two baseline cases with $n=3$ and $n=8$. 

Table~\ref{tab:nsource} indicates that the impact of the number of modality sources ($n$) on performance is subtle and dataset-dependent. Setting $n=3$ slightly improves recognition on KHOB, KhmerST, GKHST, and KHT, but slightly reduces accuracy on KH. Increasing $n$ to eight yields minor gains on KHOB at the expense of other datasets. Overall, $n=3$ and $n=5$ provide essentially equivalent performance.

\begin{table*}[t]
\centering
\caption{Character error rates (CER) for $n=3$, $n=5$, and $n=8$. * : model-specific tokenizer. Char.: character. KCC: Khmer character cluster. \textbf{Bold}: best. \emph{Italic}: second best.}
\label{tab:nsource}
\begin{tabular}{lcccccccc}
\toprule
 & Decoder & Tokenizer & KHOB & KhmerST & KH & GKST & KHT \\

\midrule
\textbf{UKTR ($n=3$, D + S\&H)} & CTC & KCC &  2.43 & 3.03 & 7.12 & 4.57 & 9.87\\
\textbf{UKTR ($n=3$, D + S\&H)} & Tr. & KCC &  \emph{2.29} & \textbf{2.12} & 5.17 & \textbf{2.99} & \textbf{5.82}\\
\midrule
\textbf{UKTR  ($n=5$, D + S\&H)} & CTC & KCC &  2.46 & 3.02 & 5.89 & 4.41 & \emph{9.52}\\
\textbf{UKTR  ($n=5$, D + S\&H)} & Tr. & KCC &  2.37 & \emph{2.19} & \textbf{4.11} & \emph{3.34} & \emph{6.10}\\
\midrule

\textbf{UKTR ($n=8$, D + S\&H)} & CTC & KCC &  2.45 & 3.09 & 6.50 & 4.79 & 9.71\\
\textbf{UKTR ($n=8$, D + S\&H)} & Tr. & KCC &  \textbf{2.07} & 2.31 & \emph{4.67} & 3.78 & 6.44\\

\bottomrule
\end{tabular}
\end{table*}

\subsubsection{Qualitative Assessment }
\begin{figure}
\centering

\begin{subfigure}[b]{0.27\textwidth}
    \centering
    {\includegraphics[width=\textwidth]{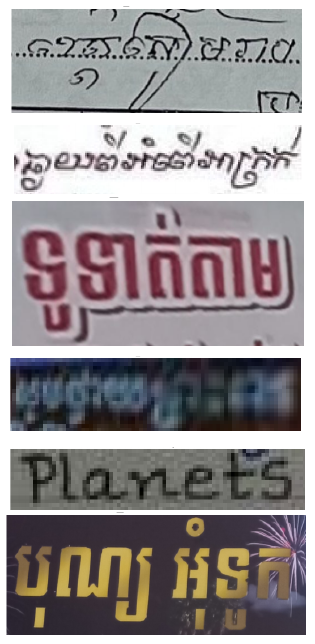}}
    \caption{image}

\end{subfigure}
\hfill
\begin{subfigure}[b]{0.20\textwidth}
    \centering
    {\includegraphics[width=\textwidth]{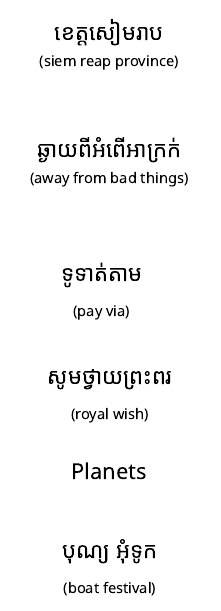}}
    \caption{ground-truth}

\end{subfigure}
\hfill
\begin{subfigure}[b]{0.21\textwidth}
    \centering
    {\includegraphics[width=\textwidth]{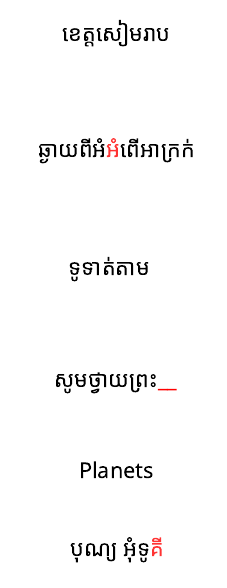}}
    \caption{CTC}

\end{subfigure}
\hfill
\begin{subfigure}[b]{0.21\textwidth}
    \centering
    {\includegraphics[width=\textwidth]{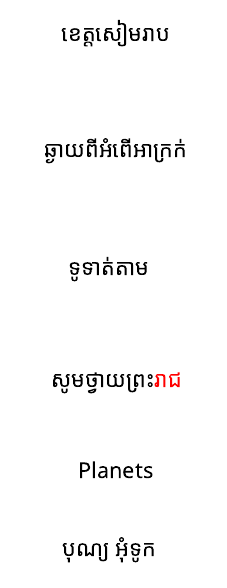}}
    \caption{Transformers}

\end{subfigure}
\caption{Sample images, ground-truth texts, and model predictions (CTC vs. Transformers).  } \label{qual_eval}
\end{figure}

Figure~\ref{qual_eval} presents qualitative comparisons between the ground-truth texts and the model predictions obtained using the CTC and Transformer decoders for several randomly selected input images. Consistent with the quantitative results, predictions from the Transformer decoder are generally more accurate than those from the CTC decoder. Except for a highly blurry input image (fourth row), the Transformer decoder successfully extracts the correct text across all modalities. In contrast, the CTC decoder tends to struggle when the input images are either severely blurred (fourth row) or contain complex backgrounds (sixth row) or with complex handwritten text (second row).

\section{Limitations and Future Work}

We identify the following limitations associated with this study and future directions as follows:

\begin{enumerate}
   \item The $\operatorname{\mathrm{ROUTER}}$ network is trained by minimizing the overall recognition loss ($\boldsymbol{l_{Total}}$) rather than any modality-specific supervision loss. Thus, future research should introduce an explicit modality supervision loss to enable the $\operatorname{\mathrm{ROUTER}}$ network to learn the labelled modality distributions.

   \item  Although the UKTR model is trained by jointly optimizing two decoders, CTC and Transformers, the choice of decoder during inference is independent of the input text image. Future research could explore a decoder router network that dynamically selects the most suitable decoder for a given image, optimizing both latency and accuracy.
\end{enumerate}

\section{Conclusions}

This paper presented a UKTR framework capable of recognizing Khmer text across multiple modalities, including document, scene, and handwritten texts. To address the limited availability of Khmer scene and handwritten training data and benchmarks, we introduced new datasets for these modalities. Experimental results demonstrate that the proposed UKTR models achieved SoTA recognition performance on the benchmark datasets.

%
%
%
%

\end{document}